%% file: acl_latex.tex
\documentclass[11pt]{article}

\usepackage[final]{acl}

\usepackage{times}
\usepackage{latexsym}

\usepackage[T1]{fontenc}

\usepackage[utf8]{inputenc}

\usepackage{microtype}

\usepackage{inconsolata}

\usepackage{graphicx}

\usepackage{enumitem}
\usepackage{colortbl}
\usepackage{amsmath}
\usepackage{multirow}
\usepackage{booktabs}
\usepackage{tabularx}
\usepackage{makecell}
\usepackage{subcaption}
\usepackage{wrapfig}
\usepackage{tcolorbox}
\usepackage{listings}
\usepackage{caption}
\usepackage{xcolor}
\tcbuselibrary{breakable}
\usepackage{appendix}

\title{After Retrieval, Before Generation: Enhancing the ‌Trustworthiness of Large Language Models in Retrieval-Augmented Generation}

\author{
Xinbang Dai$^{\dagger*}$ 
\quad Huikang Hu$^{\dagger*}$ 
\quad Yuncheng Hua$^\clubsuit$ 
\quad Jiaqi Li$^\dagger$ 
\quad Nan Hu$^\dagger$\\
\quad \textbf{Yongrui Chen}$^\dagger$
\quad \textbf{Rihui Jin}$^\dagger$
\quad \textbf{Yuxin Zhang}$^\dagger$
\quad \textbf{Yuyang Zhang}$^\diamond$\\
\quad \textbf{Xiaoguang Li}$^\diamond$ 
\quad \textbf{Lifeng Shang}$^\diamond$ 
\quad \textbf{Guilin Qi}$^\dagger$ \\
$^\dagger$Southeast University  \quad $^\clubsuit$University of New South Wales \quad $^\diamond$Noah's Ark Lab \\
\texttt{\{xbdai, huikanghu\}@seu.edu.cn} \\
\phantom{\thanks{Equal contribution.}}
}

\begin{document}
\maketitle

\begin{abstract}
Retrieval-augmented generation (RAG) is a promising paradigm, yet its trustworthiness remains a critical concern. A major vulnerability arises prior to generation: models often fail to balance parametric (internal) and retrieved (external) knowledge, particularly when the two sources conflict or are unreliable. To analyze these scenarios comprehensively, we construct the Trustworthiness Response Dataset (TRD) with 36,266 questions spanning four RAG settings. We reveal that existing approaches address isolated scenarios—prioritizing one knowledge source, naively merging both, or refusing answers—but lack a unified framework to handle different real-world conditions simultaneously. Therefore, we propose BRIDGE, a framework that dynamically determines comprehensive response strategies for large language models (LLMs). BRIDGE leverages an adaptive weighting mechanism to guide knowledge collection, followed by a decision tree to evaluate evidence and select the optimal response strategy (e.g., trusting internal/external knowledge, or refusal). Experiments show BRIDGE outperforms baselines by 5–15\% in accuracy while maintaining balanced performance across all scenarios. Our work provides an effective solution for producing trustworthy LLM responses in RAG applications.
\footnote{The code and benchmark is publicly available at \url{https://github.com/Kangkang625/BRIDGE}.}
\end{abstract}

\input{sections/1_introduction}

\input{sections/2_TRD_Construction}
\input{sections/3_BRIDGE}
\input{sections/4_experiments}

\input{sections/5_conclusion}
\input{sections/6_limitations}

\bibliography{ref}
\clearpage

\appendix

\input{sections/7_appendix}

\input{sections/8_related_works}

\end{document}

%% file: sections/1_introduction.tex
\section{Introduction}


Recent advancements in retrieval-augmented generation (RAG) have enhanced the accuracy of large language models (LLMs) in knowledge-intensive tasks~\cite{guu2020retrieval, lewis2020retrieval}. After retrieval and before response generation, LLMs often rely on two primary knowledge sources: context acquired through retrieval, referred to as \textbf{external knowledge}, and parametric knowledge stored within the LLM during training, termed \textbf{internal knowledge}. However, both sources have limitations that can undermine the trustworthiness of RAG systems. Specifically, the retrieval process may introduce irrelevant or poisoned content~\cite{chen2024benchmarking, zou2024poisonedrag}, whereas internal knowledge may be outdated or erroneous due to temporal limitations or noise in the training data~\cite{weidinger2021ethical, longpre2024pretrainer}.

\begin{figure*}[t]
  \centering
  \includegraphics[width=0.94\textwidth]{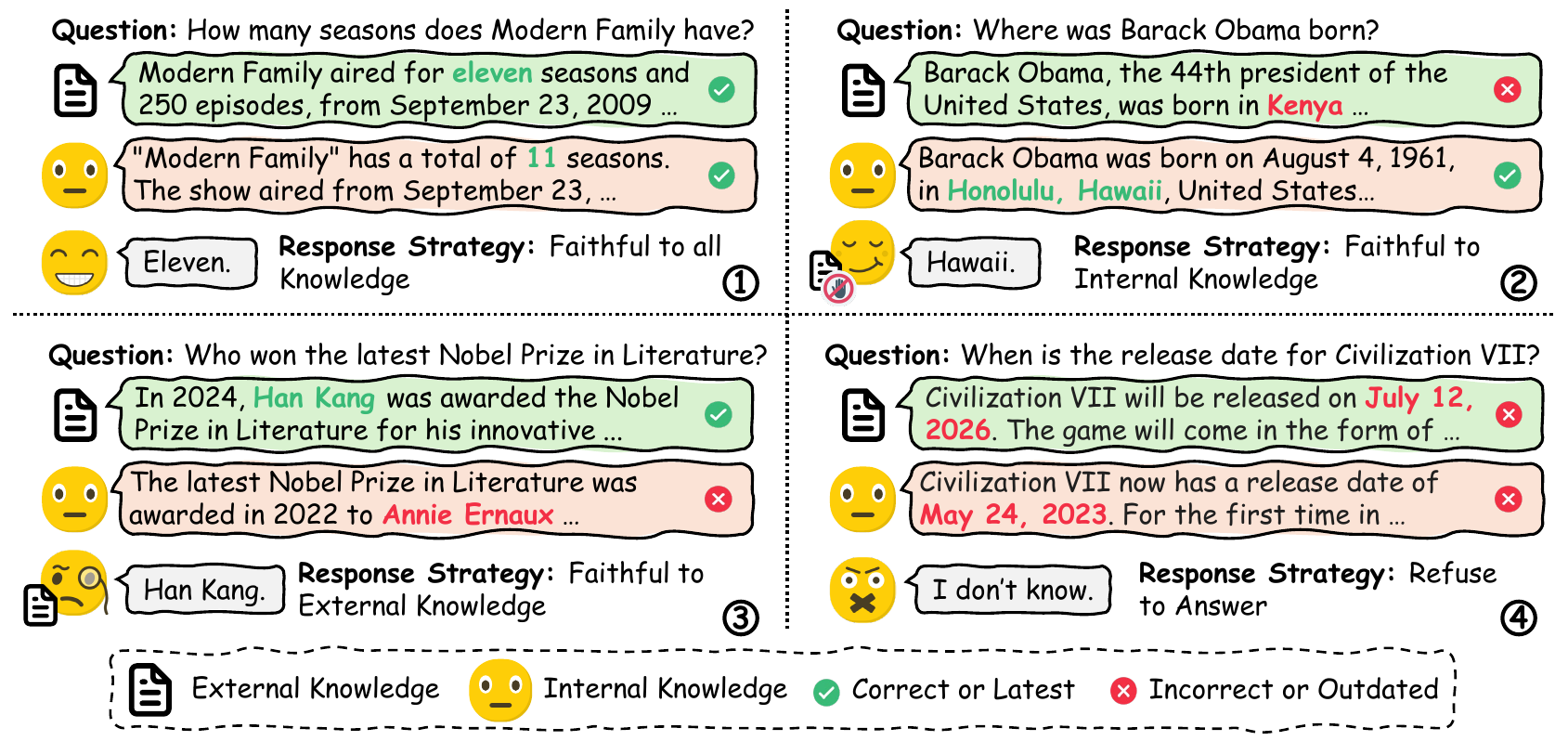}
  \caption{Four scenarios in real-world RAG.}
  \label{fig1}
\end{figure*}

In real-world RAG applications, limitations in internal and external knowledge sources can lead to four distinct scenarios~\cite{wang2024astute,zhang2024evaluating} (as illustrated in Figure~\ref{fig1}).
Ideally, LLMs should adopt scenario-specific strategies:
(1) When both knowledge sources are consistent and correct, LLMs should base their responses on all available knowledge.
(2) If external knowledge contains errors while internal knowledge is reliable, LLMs should rely on their internal knowledge.
(3) Conversely, when internal knowledge is outdated or incorrect but the retrieved context is valid, LLMs should anchor responses to external evidence.
(4) When neither source contains sufficient or credible information, LLMs should refuse to answer.

Unfortunately, existing research typically addresses these scenarios in isolation. Some studies enforce alignment of LLM responses with a single source (either internal or external)~\cite{pan2023risk, chen2024universal, li2022large, weller2022defending, zhou2023context}, whereas others integrate both sources but overlook the need for refusal mechanisms (Scenario 4)~\cite{shi2024trusting, asai2023self, xiang2024certifiably, wang2024astute, wei2025instructrag, zhou2025trustrag}. Conversely, safety-focused works discuss refusal but neglect effective knowledge integration~\cite{cao2023learn, song2025measuring}. As a result, existing frameworks do not simultaneously consider all four scenarios required for trustworthy RAG.

To systematically evaluate LLM response strategies across these four scenarios, we introduce the Trustworthiness Response Dataset (TRD), comprising 36,266 multiple-choice questions. TRD covers four categories, each aligned with a desired response strategy: (1) \emph{Faithful to All Knowledge (FA)}: both internal and external knowledge support the correct answer, so models should base responses on all available knowledge; (2) \emph{Faithful to Internal Knowledge (FI)}: external context is deliberately poisoned while internal knowledge remains reliable, so models should rely on internal knowledge; (3) \emph{Faithful to External Knowledge (FE)}: questions require up-to-date facts beyond the model's pre-training cutoff dates, so internal knowledge is insufficient and models must rely on external evidence; and (4) \emph{Refuse to Answer (RA)}: both internal and external knowledge are unreliable, so models should refuse to answer.

Our evaluation on TRD shows that when the model is simultaneously exposed to two potentially imperfect knowledge sources, it may adopt unreasonable response strategies. Moreover, existing LLM families and RAG baselines exhibit refusal behavior in fewer than 10\% of cases (as shown in Table~\ref{tab1}).
Human behavioral patterns offer inspiration: humans verify known facts with minimal retrieval~\cite{fu2006suboptimal}, deeply evaluate external sources for uncertain or time-sensitive questions~\cite{rieh2002judgment}, and readily admit ignorance when evidence is lacking~\cite{gigerenzer2010rationality}.

Therefore, we propose \textbf{B}iased \textbf{R}etrieval an\textbf{D} \textbf{G}eneration \textbf{E}valuation (BRIDGE), a framework that enhances the trustworthiness of LLMs in RAG systems. Unlike prior work that either enforces strict alignment with a single knowledge source (hard bias) or treats all sources equally (no bias), BRIDGE introduces \textbf{soft bias}: an adaptive weighting mechanism where an \textbf{Allocator} module determines whether to conduct deeper internal knowledge generation (for fact verification) or prioritize broad external knowledge retrieval (for consistency evaluation). Based on the TRD training data, Allocator supports both in-context learning (ICL) and group relative policy optimization (GRPO) paradigms.
Subsequently, based on the weighted knowledge sources, BRIDGE computes matching scores and evaluates consistency through a \textbf{Maximum Soft-bias Decision Tree}—an interpretable decision tree enabling fine-grained response strategy selection (e.g., trusting internal knowledge, following external evidence, or refusing to answer). Additionally, it incorporates a reflection strategy to rectify decision errors during inference. In summary, our contributions are as follows:
\begin{itemize}
\item We construct TRD, a comprehensive benchmark designed to evaluate LLM response strategies across diverse RAG scenarios.

\item We propose BRIDGE, a unified framework that addresses diverse RAG scenarios.

\item Experiments demonstrate that BRIDGE achieves superior and more balanced performance compared to existing methods, not only on TRD but also on established benchmarks targeting individual RAG scenarios.
\end{itemize}

%% file: sections/2_TRD_Construction.tex
\section{TRD Construction}

This section details the construction process of TRD. We formalize our task scenario as follows: given a natural language question $q$, the system retrieves external knowledge $K_{ext}$ and generates explicit internal knowledge $K_{int}$ by answering $q$ without accessing $K_{ext}$. Each question in TRD is associated with four options: one correct option and three incorrect options. One option is "I don't know," allowing the model to abstain from answering. The dataset comprises 29,012 questions for training, 3,627 for validation, and 3,627 for testing.

\subsection{Data Preparation}
We use two datasets to build TRD’s multiple-choice questions.
(1) NQ~\cite{kwiatkowski2019natural}: An open-domain QA dataset with 315,203 QA pairs, each paired with a long answer (Wikipedia passage) and one or more short answer entities. We use NQ to generate questions for the \emph{FA} and \emph{FI} categories.
(2) TAQA~\cite{zhao2024set}: A dataset of 20,148 time-sensitive QA pairs from Wikipedia tables, where answers change over time. We use TAQA to construct part of the \emph{FE} and \emph{RA} questions.

\subsection{Question Generation}
\label{question_generation}

\textbf{Faithful to All Knowledge (FA).}  
From the NQ training set, we designate the short answer as the correct option for each question. Using a popularity-based entity substitution method~\cite{longpre2021entity}, we generate two incorrect answers from Wikidata that share the same entity type as the short answer. For each question, the corresponding long answer serves as the external knowledge $K_{ext}$. To obtain the internal knowledge for all question types (\emph{FA}, \emph{FI}, \emph{FE}, \emph{RA}), we prompt three backbone models ({\tt GPT-3.5-turbo}, {\tt Qwen-72B}, and {\tt Llama-3-8B-Instruct}) to answer each question directly. We regard the generated responses as explicit representations of the models' internal knowledge.

\textbf{Faithful to Internal Knowledge (FI).}  
Questions of type \emph{FI} are derived from the NQ validation set. The generation procedure for answer options follows that of \emph{FA}. To construct misleading external knowledge, we replace the correct entity in the long answer with one of the incorrect entities from the distractor options, resulting in a corrupted $K_{ext}$ that introduces factual conflict while preserving textual plausibility.

\textbf{Faithful to External Knowledge (FE).}
To simulate scenarios where LLMs must rely solely on external knowledge due to outdated or absent internal knowledge, we define two distinct types: 

(1) \emph{Evolution}: This category covers questions that require updated information about existing facts and events (e.g., \emph{Who won the \textbf{latest} Nobel Prize in Literature?}). LLMs with a training data cutoff (in our paper, February 2023) are unable to track such updates.
Following the methodology described in the TAQA dataset paper, we update the short answers and their corresponding contexts using a Wikipedia dump dated February 1, 2025. The context extracted from Wikipedia serves as external knowledge $K_{ext}$, with the updated short answers treated as correct options. The original answers from the TAQA dataset (dated 2020 and 2021) are retained as incorrect distractors.

(2) \emph{Perpetuation}:
This category addresses new entities or events that have emerged after the LLM's knowledge cutoff (e.g., \emph{What is the CPU of iPhone 16?}). For many LLMs whose training data ends in 2023, such information remains unknown. First, we execute SPARQL\footnote{https://www.wikidata.org/wiki/Wikidata:SPARQL\_tutorial} queries on Wikidata to extract timestamped triples created after February 2023 (e.g., <\emph{iPhone 16, hasCPU, A18}>). For each of these triples, we retrieve textual paragraphs from Wikipedia that contain both the subject and object entities as external knowledge. We then mask the object entity of the triple and prompt {\tt GPT-4-turbo} to generate a question for which the masked entity serves as the correct answer.
In addition, we retrieve two incorrect distractor options using the substitution method~\cite{longpre2021entity}. It is worth noting that when relying on internal knowledge, the model is unaware of these new facts; consequently, its responses typically consist of hallucinated content or explicit admissions of ignorance. Finally, we split the resulting dataset evenly: one half is used as \emph{FE} data, and the other half is used to construct subsequent \emph{RA} data.

\textbf{Refuse to Answer (RA).}
We corrupt the external knowledge text by replacing the correct answer with a distractor entity (e.g., changing \emph{A18} to \emph{A17} in the iPhone 16 example). This ensures that no credible evidence exists in any knowledge source, creating a scenario in which the model should ideally refuse to answer. Finally, we set the correct answer to "I don't know."

\begin{figure*}[t]
  \centering
  \includegraphics[width=\textwidth]{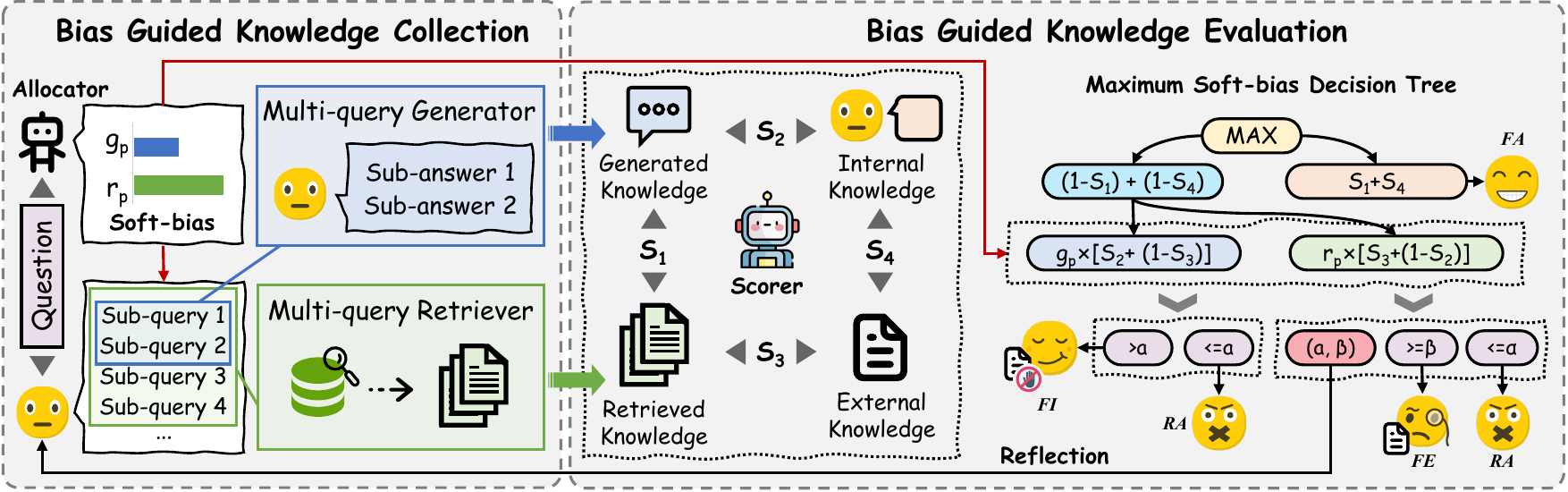}
  \caption{Overview of BRIDGE.}
  \label{fig2}
\end{figure*}

\subsection{Question Validation}
To validate the questions in our TRD benchmark, we use the backbone LLMs employed in this study for verification and filtering. This validation process adheres to a strict set of criteria for each question category. (1) For \emph{FA}, a question is retained only if the models can answer it correctly when given only the internal knowledge source or only the external knowledge source. (2) For \emph{FI}, a question is retained only if the models answer it correctly using internal knowledge alone but, when provided with erroneous external information, are misled into selecting the target incorrect answer. (3) For \emph{FE}, a question is retained if the models fail to produce the correct answer when relying exclusively on internal knowledge. (4) For \emph{RA}, a question is retained if the models cannot derive the correct answer from either knowledge source. For additional details on TRD construction and statistics, please refer to Appendix~\ref{trd_appendix}.


%% file: sections/3_BRIDGE.tex
\section{Biased Retrieval and Generation Evaluation}

As illustrated in Figure~\ref{fig2}, BRIDGE processes the user question $q$ in two stages:
(1) \emph{Bias-Guided Knowledge Collection}: The Allocator first estimates how much the question $q$ relies on internal versus external knowledge, producing the generation and retrieval dependency probabilities ($g_p$ and $r_p$, where $r_p + g_p = 100\%$) (\S~\ref{allocator}). The system then generates $n$ sub-queries that capture key information from $q$, and allocates them to the Multi-query Generator and the Multi-query Retriever according to $g_p$ and $r_p$, yielding generated knowledge $K_{gen}$ and retrieved knowledge $K_{ret}$ (\S~\ref{knowledge_collection}).
(2) \emph{Bias-Guided Knowledge Evaluation}: The framework evaluates four knowledge sources: self-generated internal knowledge $K_{int}$ , initially retrieved external knowledge $K_{ext}$, and the previously obtained $K_{gen}$ and $K_{ret}$. We employ the {\tt BGE-m3} encoder~\cite{chen2024bge} to compute multi-granularity similarity scores across knowledge types (\S\ref{scorer}). Then, a Maximum Soft-bias Decision Tree uses these scores to determine the optimal response strategy. Moreover, if potential errors are detected, a reflection module triggers re-execution for refinement (\S\ref{search_tree}).

\subsection{Bias-Guided Knowledge Collection}

\subsubsection{Soft Bias Allocator}
\label{allocator}

Determining soft bias (i.e., $r_p$ and $g_p$) for a single question presents significant challenges. For instance, the question \emph{What is the CPU of the iPhone 16?} contains no time-sensitive keywords, requiring the LLM to assign dependency probabilities by analyzing the question's semantics and assessing its internal knowledge boundaries. To enhance this capability, we leverage TRD's training data to construct annotated examples with analysis paths. We develop two Allocator paradigms: (1) a GRPO-optimized {\tt Llama3-8B-Instruct} (pretrained cutoff: Feb 2023), and (2) an ICL-based approach.
 
\textbf{Soft Bias Analysis Data Construction.} We construct data through three steps:  
(1) We assign the retrieval dependency probability $r_p^{hard}$ and the generation dependency probability $g_p^{hard}$ according to the scenario type in TRD. These hard bias values serve as the ground-truth labels for soft bias learning.
For TRD's \emph{FA/FI} categories, we set $r_p^{hard}=0\%$, $g_p^{hard}=100\%$; for \emph{RA} and \emph{FE}, $g_p^{hard}=0\%$, $r_p^{hard}=100\%$.  
(2) Using {\tt DeepSeek-R1} and {\tt o3-mini} as reasoning models, we instruct them to generate an analysis path $a$, along with soft bias probabilities $r_p^{soft}$ and $g_p^{soft}$ constrained to the range $[10\%, 90\%]$, under explicit directional guidance derived from the hard bias labels (e.g., $g_p^{soft} > r_p^{soft}$ for \emph{FA/FI} queries).
(3) We filter for high-quality data by selecting the generated outputs that exhibit the lowest root mean squared error between the soft ($r_p^{soft}, g_p^{soft}$) and hard ($r_p^{hard}, g_p^{hard}$) bias labels. 

\textbf{Allocator (GRPO).} 
We employ GRPO to fine-tune our Allocator, using a composite reward function to optimize {\tt Llama3-8B-Instruct}. We instruct the model to output a tuple $\gamma = (a_{pred}, r_p^{pred}, g_p^{pred})$, comprising the analysis path $a_{pred}$, predicted retrieval dependency probability $r_p^{pred}$, and generation dependency probability $g_p^{pred}$. 
The training objective consists of four distinct reward components: \emph{Direction Reward}, \emph{Format Reward}, \emph{Sum Reward}, and \emph{Analysis Quality Reward}. The specific functions and designs of these rewards are detailed in the Appendix~\ref{grpo_train_details}.

\textbf{Allocator (ICL).} We also implement an ICL paradigm by retrieving the top-k similar training examples for each TRD. Instead of relying on soft-bias outputs and analysis paths from reasoning models, we directly convert the gold hard-bias presets into soft-bias labels (e.g., 100\%→90\%, 0\%→10\%) to construct pseudo soft-bias demonstrations. This ensures that BRIDGE’s performance improvements are not attributable to external reasoning model abilities. We discuss the analysis in \S~\ref{main_results} and provide implementation details of the Allocator (GRPO and ICL) in Appendix~\ref{allocator_appendix}.

\subsubsection{Knowledge Collection}
\label{knowledge_collection}

To capture broader knowledge, we prompt the LLM to generate $n$ sub-queries that encapsulate the key information in $q$.  These sub-queries act as "probes," enabling both the retriever and the LLM to explore a wider range of knowledge within their respective domains.
Specifically, we allocate sub-queries proportionally to each knowledge source based on the Allocator’s soft bias probabilities. Given $n$ sub-queries, the numbers assigned to retrieval ($s_r$) and generation ($s_g$) are computed as $s_r = \lceil r_p^{pred} \times n \rceil$ and $s_g = \lceil g_p^{pred} \times n \rceil$, respectively. The first $s_r$ sub-queries are processed by the Multi-query Retriever, which aggregates external documents to form the retrieved knowledge$K_{ret}$, while the first $s_g$ sub-queries are answered by the Multi-query Generator, which produces the generated knowledge $K_{gen}$. This adaptive allocation strategy enhances the overall effectiveness of knowledge acquisition.

\subsection{Bias-Guided Knowledge Evaluation}

\subsubsection{Knowledge Scorer}
\label{scorer}

In this stage, we compute three similarity measures between knowledge sources using the BGE Scorer~\cite{chen2024bge}: sparse matching to capture lexical overlap, dense matching to assess semantic similarity, and ColBERT-based matching~\cite{khattab2020colbert} to capture token-level correlations. The respective weights are determined by grid search on the TRD validation set (details are provided in Appendix~\ref{grid_search}). This multi-granularity strategy provides a more comprehensive evaluation of knowledge relevance and demonstrates broad applicability. As validated in \S~\ref{main_results}, all weights and hyperparameters, tuned exclusively on TRD, achieve balanced performance when applied directly to other datasets without further modification, thereby confirming their robustness.

\subsubsection{Maximum Soft-bias Decision Tree}
\label{search_tree}

After obtaining four matching scores using the Scorer, we perform three sequential checks:

\textbf{Consistency Detection.} If $S_1 + S_4 > (1-S_1) + (1-S_4)$ (implying high consistency between the two knowledge sources), the system triggers \emph{FA}, and the LLM integrates all knowledge.

\textbf{Confidence Calculation.} 
We compute the trustworthiness scores for the retriever and the LLM as follows: $T_{Ret} = r_p^{pred}[S_3 + (1-S_2)]$, $T_{LLM} = g_p^{pred}[S_2 + (1-S_3)]$. This formulation reflects two key insights: (1) a higher dependency probability ($r_p^{pred}$ or $g_p^{pred}$) amplifies the influence of the corresponding knowledge matches, and (2) inconsistencies in one knowledge source reduce its trustworthiness score, while consistency in the other source may further attenuate the first source’s score. For example, poor alignment between $K_{gen}$ and $K_{int}$ ($S_2$→0) and good alignment between $K_{ret}$ and $K_{ext}$ ($S_3$→1) both contribute to lowering $T_{LLM}$.

\textbf{Threshold-based Decision.} As shown in Figure~\ref{fig2}, we employ two thresholds ($\alpha$, $\beta$) determined by grid search on the TRD validation set (see Appendix~\ref{grid_search}). For cases where $T_{Ret} < T_{LLM}$, the system adopts \emph{FI} when $T_{LLM} > \alpha$, otherwise triggering \emph{RA}. When $T_{Ret} \geq T_{LLM}$, the system's response is determined by the value of $T_{Ret}$: it triggers \emph{FE} if $T_{Ret} \geq \beta$, initiates \emph{RA} if $T_{Ret} < \alpha$, or invokes the Reflection mechanism~\cite{madaan2023self} to adjust sub-queries for intermediate cases ($\alpha \leq T_{Ret} < \beta$). After three unsuccessful cycles, the system defaults to \emph{RA} as a conservative response.
The asymmetric thresholds ($\beta > \alpha$) mean the retriever knowledge source requires higher trustworthiness due to potential retrieval noise.
Finally, our framework dynamically adjusts response strategies based on predicted interpretable decisions: \emph{FA} employs all knowledge; \emph{FI} relies solely on $K_{int}$ and $K_{gen}$; \emph{FE} prioritizes $K_{ext}$ and $K_{ret}$; and \emph{RA} responds with "I don’t know". All system prompts are detailed in Appendix~\ref{bridge_system_prompts}.

\begin{table*}[t]
\centering
\resizebox{\textwidth}{!}{%
\begin{tabular}{lcccccccccccc}
\toprule

\multirow{3}{*}{Methods} & \multicolumn{4}{c}{TRD} & \multicolumn{2}{c}{RQA} & \multicolumn{2}{c}{HQA\textsubscript{P}} & \multicolumn{4}{c}{\sc ConflictBank}\\ 

\cmidrule(r){2-5} \cmidrule(r){6-7} \cmidrule(r){8-9} \cmidrule(r){10-13} 

 & \multicolumn{2}{c}{Simu} & \multicolumn{2}{c}{Real} & \multicolumn{1}{c}{Simu} & \multicolumn{1}{c}{Real} & \multicolumn{1}{c}{Simu} & \multicolumn{1}{c}{Real} & \multicolumn{2}{c}{Simu} & \multicolumn{2}{c}{Real}\\ 

 \cmidrule(r){2-3} \cmidrule(r){4-5} \cmidrule(r){6-6} \cmidrule(r){7-7} \cmidrule(r){8-8} \cmidrule(r){9-9} \cmidrule(r){10-11} \cmidrule(r){12-13}

 & Acc(\%) & RR(\%) & Acc(\%) & RR(\%) & EM(\%) & EM(\%) & EM(\%) & EM(\%) & Acc(\%) & RR(\%) & Acc(\%) & RR(\%) \\ 

\midrule
 
\multicolumn{13}{c}{\cellcolor{gray!25}\emph{Based on} {\tt GPT-3.5-turbo}} \\ 

\midrule
Vanilla & 53.71 & 5.81 & 55.80 & 14.61 & 72.52 & 29.41 & 44.40 & 69.23 & 52.04 & 0.20 & 38.81 & 38.24 \\
OPIN~\cite{zhou2023context} & 63.93 & 3.34 & 54.62 & 16.44 & 76.25 & 23.75 & 23.47 & 54.55 & 48.03 & 2.00 & 37.02 & 40.71 \\
USC~\cite{chen2024universal} & 65.53 & 3.01 & 50.41 & 15.77 & 26.25 & 25.71 & 70.70 & 73.47 & 73.46 & 1.08 & 40.21 & 29.87 \\
RobustRAG~\cite{xiang2024certifiably} & 62.50 & 6.69 & 56.94 & 14.69 & 65.82 & 29.11 & 64.00 & 73.00 & 58.40 & 3.04 & 39.39 & 25.25 \\
InstructRAG\textsubscript{ICL}~\cite{wei2025instructrag} & 65.55 & 4.29 & 57.11 & 20.84 & 74.36 & 28.75 & 57.30 & 54.55 & 57.45 & 1.71 & 37.71 & 40.02 \\
AstuteRAG~\cite{wang2024astute} & 64.26 & 2.81 & 58.60 & 14.42 & 74.03 & 26.25 & 43.43 & 71.12 & 57.45 & 1.66 & 31.15 & \textbf{46.28} \\
TrustRAG~\cite{zhou2025trustrag} & 66.39 & 3.42 & 55.53 & 15.03 & \textbf{78.48} & 32.50 & 40.00 & 74.85 & 52.01 & 2.50 & 27.66 & 42.55 \\
BRIDGE\textsubscript{ICL} & 72.32 & 31.40 & 68.72 & \textbf{32.43} & 76.92 & \textbf{33.66} & 70.97 & 79.69 & 61.86 & \textbf{38.14} & 76.15 & 28.42 \\ 
BRIDGE\textsubscript{GRPO} & \textbf{77.90} & \textbf{33.33} & \textbf{75.66} & 22.71 & \textbf{78.48} & 33.38 & \textbf{72.16} & \textbf{81.23} & \textbf{75.31} & 24.69 & \textbf{77.14} & 24.61 \\ 

\midrule

\multicolumn{13}{c}{\cellcolor{gray!25}\emph{Based on} {\tt Qwen 72B}} \\ 

\midrule

Vanilla & 63.19 & 9.50 & 52.20 & 14.82 & 70.25 & 30.12 & 61.61 & 71.31 & 57.83 & 0.61 & 51.37& 31.72 \\
OPIN~\cite{zhou2023context} & 63.40 & 3.41 & 58.22 & \textbf{54.53} & 75.95 & 31.22 & 32.65 & 59.21 & 42.22 & 4.44 & 23.02 & \textbf{57.21} \\
USC~\cite{chen2024universal} & 64.20 & 4.80 & 53.51 & 8.82 & 28.75 & 26.25 & \textbf{76.00} & 86.97 & \textbf{70.74} & 2.04 & 62.41 & 16.32 \\
RobustRAG~\cite{xiang2024certifiably} & 64.33 & 7.28 & 54.07 & 9.76 & 74.25 & 26.25 & 65.00 & 72.81 & 59.72 & 3.04 & 55.41 & 24.71 \\
InstructRAG\textsubscript{ICL}~\cite{wei2025instructrag} & 64.69 & 4.71 &  53.27 & 9.66 & 73.75 & 25.17 & 67.32 & 72.61 & 56.80 & 1.08 & 53.16 & 17.48 \\
AstuteRAG~\cite{wang2024astute} & 66.29 & 3.60 & 55.83 & 7.61 & 75.00 & 23.08 & 66.35 & 86.62 & 66.89 & 1.73 & 52.17 & 19.26 \\
TrustRAG~\cite{zhou2025trustrag} & 66.41 & 3.25 & 54.31 & 5.41 & 74.68 & 24.05 & 67.47 & 83.88 & 63.75 & 1.25 & 53.81 & 16.48 \\
BRIDGE\textsubscript{ICL} & 74.20 & \textbf{39.56} & 58.78 & 27.32 & 76.25 & 32.11 & 72.15 & 85.32 & 67.35 & \textbf{32.65} & 74.18 & 23.81 \\ 
BRIDGE\textsubscript{GRPO} & \textbf{85.48} & 32.34 & \textbf{76.86} & 27.07 & \textbf{84.06} &  \textbf{36.71} & 75.00 & \textbf{87.61} & 69.39 & 30.61 & \textbf{75.27} & 16.82 \\ 

\midrule

\multicolumn{13}{c}{\cellcolor{gray!25}\emph{Based on} {\tt Llama3 8B-Instruct}} \\ 

\midrule
Vanilla & 42.20 & 1.80 & 51.80 & 19.80 & 68.75 & 25.00 & 57.57 & 64.00 & 46.00 & 0.81 & 33.00 & 22.00 \\
SelfRAG~\cite{asai2023self} & 47.01 & 14.97 & 48.92 & 18.26 & 61.25 & 27.12& 55.42 & 68.31 & 61.25 & 8.18 & 36.25 &15.21 \\
CAD~\cite{shi2024trusting} & 24.40 & 1.50 & 45.21 & 19.15& 55.00 & 20.00 & 35.00 & 58.13& 39.20 & 1.95 & 20.13 & \textbf{27.75} \\
InstructRAG\textsubscript{FT}~\cite{wei2025instructrag} & 51.81 & 13.43 & 53.61 & 15.32 & 69.30 & 23.75 & 70.70 & 41.07 & 60.78 & 15.00 & 54.28 & 16.07 \\
{\sc Trust-Align}~\cite{song2025measuring} & 53.05 & 18.11 & 48.62 & 21.61 & 72.50 & 16.25 & 69.30 & 33.00 & 65.61 & 5.10 & 20.80 & 21.45 \\
BRIDGE\textsubscript{ICL} & 62.94 & \textbf{23.86} & 61.52 & \textbf{33.67} & 78.75 & 32.52 & 71.72 & 73.11 & 59.60 & \textbf{31.46} & 41.21 & 21.66 \\ 
BRIDGE\textsubscript{GRPO} & \textbf{73.33} & 19.30 & \textbf{72.02} & 23.68 & \textbf{80.00} & \textbf{32.91} & \textbf{78.72} & \textbf{90.42} & \textbf{71.72} & 28.28 & \textbf{62.42} & 18.64 \\ 
\bottomrule
\end{tabular}%
}
\caption{Results are reported in terms of accuracy (Acc), rejection rate (RR, regardless of whether the question is answerable), and exact match score (EM)~\cite{rajpurkar2016squad}, with the best results highlighted in \textbf{bold}.}
\label{tab1}
\end{table*}

%% file: sections/4_experiments.tex
\section{Experiments}

\subsection{Experimental Setups}
\textbf{Datasets.} We evaluate all methods on four datasets.
(1) \emph{TRD}: This is the test set that we constructed. 
(2) \emph{RealtimeQA (RQA)}~\cite{kasai2024realtime}: A dynamic news QA platform. We select 468 questions that emerged after October 2023 and provide their corresponding evidence as external knowledge. This dataset aligns with the \emph{FE} scenario.  
(3) \emph{HotpotQA\textsubscript{Poisoned} (HQA\textsubscript{P})}\cite{yang2018hotpotqa}: We use the version from RobustRAG~\cite{xiang2024certifiably}, where external knowledge is corrupted by PoisonedRAG~\cite{zou2024poisonedrag}. This dataset corresponds to the \emph{FI} scenario.
(4) \emph{\sc ConflictBank}\cite{su2024conflictbank}: This is a multiple-choice dataset containing temporally misaligned external knowledge fabricated from Wikidata entries with future timestamps. We designate the fabricated temporal misalignment entries as external knowledge. Then, we randomly sample 1,000 questions from the dataset to evaluate the baselines under the \emph{FI} scenario.
The internal knowledge generation for all LLMs follows \S~\ref{question_generation}.

\textbf{Baselines.} Our baselines contain various aspects of LLM trustworthiness in RAG systems. Prompt-based methods include OPIN~\cite{zhou2023context}, USC~\cite{chen2024universal}, RobustRAG~\cite{xiang2024certifiably}, InstructRAG\textsubscript{ICL}~\cite{wei2025instructrag}, AstuteRAG~\cite{wang2024astute}, and TrustRAG~\cite{zhou2025trustrag}. GPT does not support deterministic decoding; hence, we report results averaged from three runs. Fine-tuning-based methods encompass SelfRAG~\cite{asai2023self}, CAD~\cite{shi2024trusting}, InstructRAG\textsubscript{FT}~\cite{wei2025instructrag} and {\sc Trust-Align}~\cite{song2025measuring}. 

\textbf{RAG Setup.} We evaluate RAG baselines using three backbone LLMs with knowledge cutoff dates prior to February 2023. For rigorous quantitative assessment, we first conduct experiments in a controlled setting where external knowledge is explicitly provided, simulating the initial output of a retriever; this setup is referred to as \textbf{Simu}. To further assess the practical applicability of these methods, we implement a realistic scenario termed \textbf{Real}, in which the retriever queries a Wikipedia dump using the user's question and dynamically retrieves the top-3 documents as external knowledge $K_{\text{ext}}$. To ensure a fair comparison, all methods employ the same retriever configuration and access identical Wikipedia dumps, with a knowledge cutoff date of February 1, 2025. Further details on the RAG setup are provided in Appendix~\ref{experiment_appendix}.

\subsection{Main Results}
\label{main_results}

As shown in Table~\ref{tab1}, compared to vanilla RAG, OPIN, which prioritizes external knowledge, achieves notable gains on RQA,  but suffers performance drops on HQA\textsubscript{P} and {\sc ConflictBank}. Conversely, USC, which enforces internal knowledge consistency, excels on HQA\textsubscript{P} and {\sc ConflictBank} but deteriorates sharply on RQA. Knowledge integration methods (RobustRAG, InstructRAG, AstuteRAG, and TrustRAG) demonstrate consistent improvements over vanilla RAG on TRD, underscoring the effectiveness of combining different knowledge sources. However, their lower RR scores indicate that their response strategies for refusing to answer are insufficiently considered.

CAD, which lacks mechanisms to resolve knowledge conflicts or reject uncertain questions, underperforms across most scenarios. In contrast, SelfRAG, InstructRAG\textsubscript{FT}, and {\sc Trust-Align}, which dynamically select knowledge sources or validate knowledge boundaries, achieve significantly higher RR rates. Nevertheless, their simple knowledge integration strategy limits broader applicability in all scenarios.
As described in \S\ref{allocator}, BRIDGE\textsubscript{ICL}, despite relying solely on templated soft bias demonstrations (without using data from large-scale reasoning models), outperforms baselines, which validates that the benefits arise from the soft-bias design rather than from synthetic data produced by larger reasoning models. BRIDGE\textsubscript{GRPO} further excels with its adaptive bias allocation, achieving balanced performance across all scenarios.
Notably, on the {\sc ConflictBank} dataset, BRIDGE achieves a combined Acc and RR of 100\% when using {\tt GPT-3.5-Turbo} and {\tt Qwen72B}. This demonstrates our model's highly reliable response strategy: when unable to provide correct answers using its internal knowledge, it chooses to abstain answering rather than risking incorrect outputs.

\subsection{Detailed Results} 
\begin{figure}[htbp]
  \centering
  \includegraphics[width=\columnwidth]{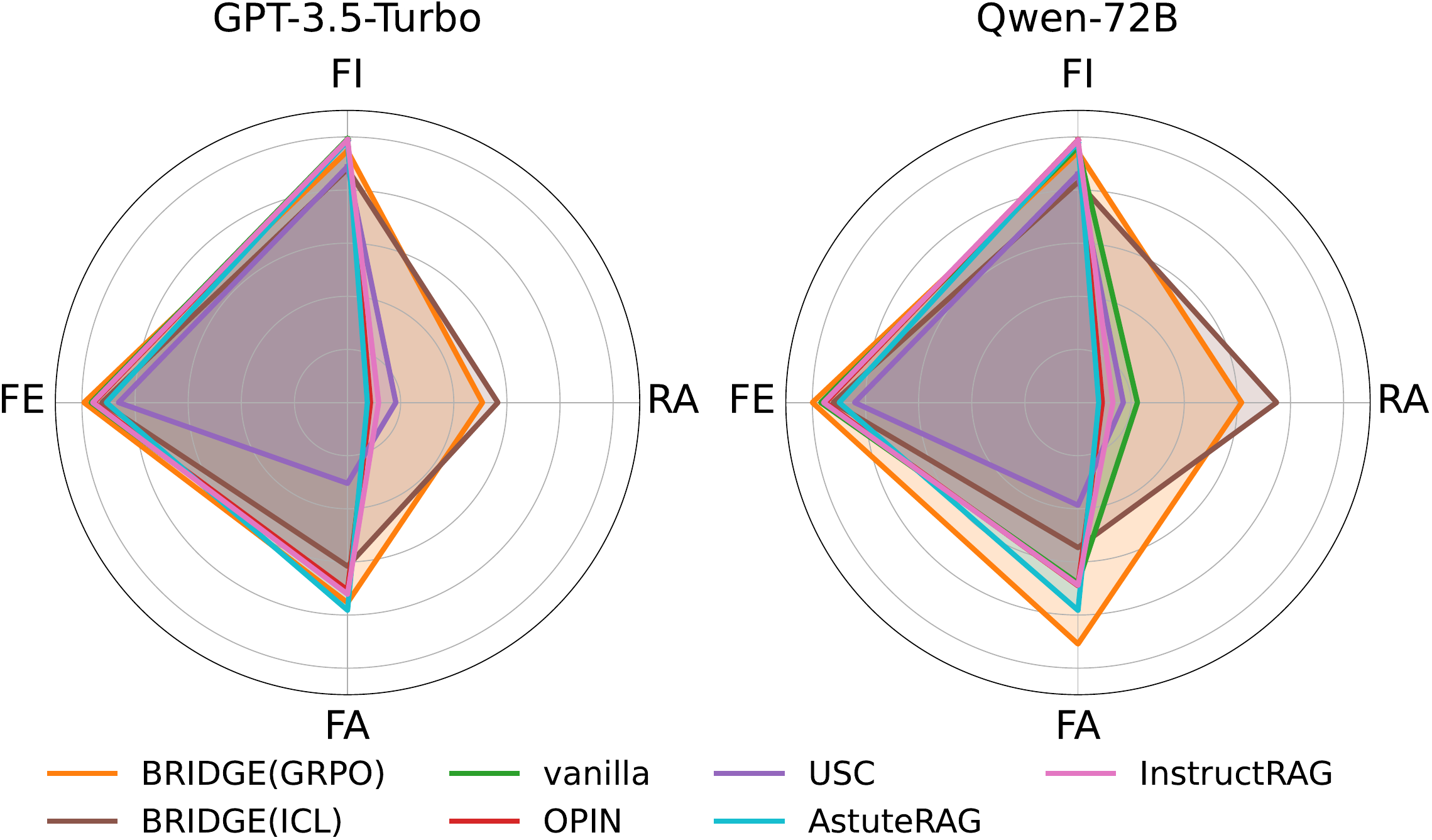}
  \caption{Acc in different RAG scenarios.}
  \label{fig3}
\end{figure}

As shown in Figure~\ref{fig3}, we compare the response accuracy of different methods across various RAG scenarios in TRD. BRIDGE demonstrates balanced capabilities across all scenarios, achieving performance comparable to state-of-the-art baselines on \emph{FA}, \emph{FI}, and \emph{FE} cases while attaining optimal performance on \emph{RA} scenarios. 

\begin{figure}[htbp]
  \centering
  \includegraphics[width=0.95\columnwidth]{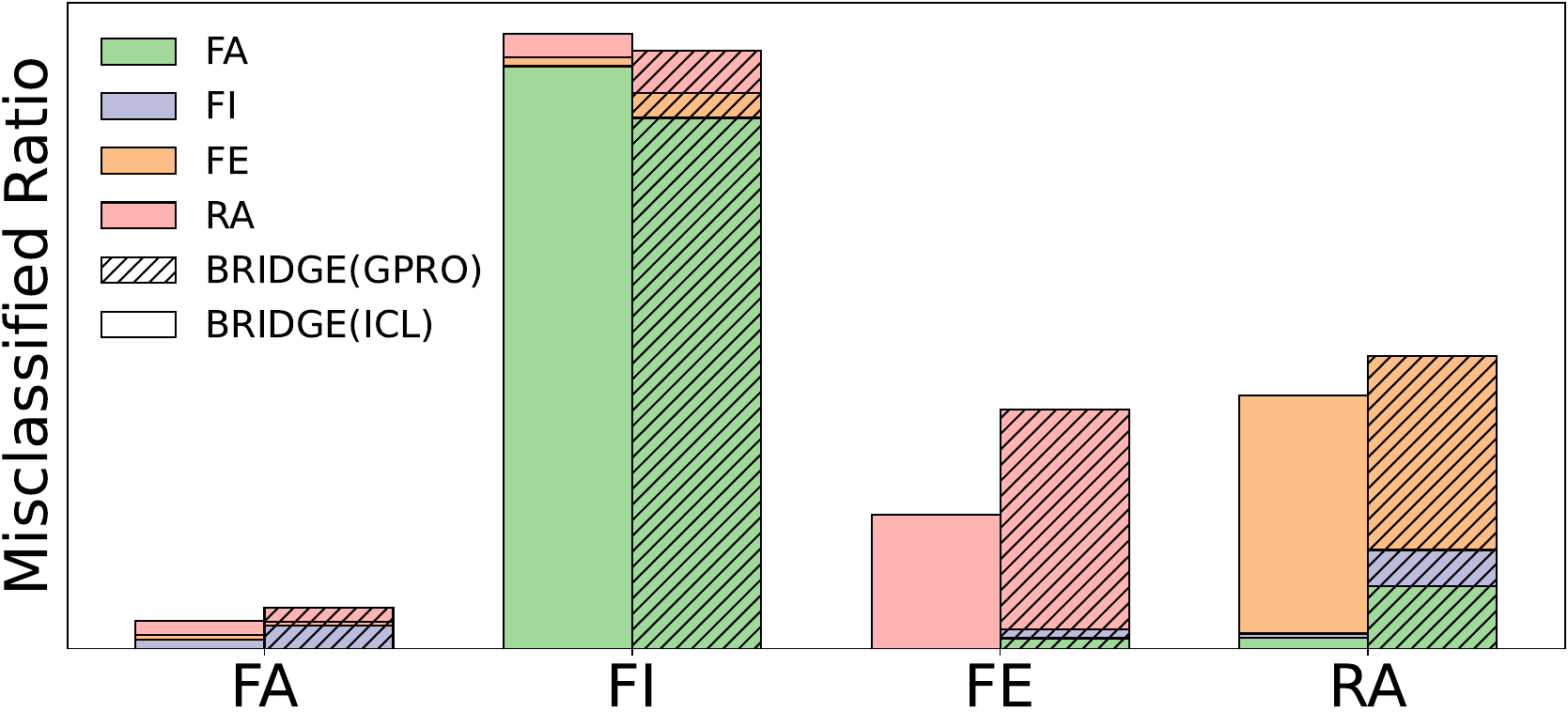}
  \caption{Misclassification distribution of BRIDGE based on {\tt GPT-3.5-Turbo}.}
  \label{fig4}
\end{figure}

Figure~\ref{fig4} presents the error response strategy distribution of BRIDGE based on {\tt GPT-3.5-Turbo}. The results reveal that \emph{FA} exhibits the lowest error probability, whereas \emph{FI} shows the highest. Notably, most \emph{FI} cases are misclassified as \emph{FA}, preserving the model's ability to access the internal knowledge source. For \emph{FE} misclassifications, the model typically adopts a conservative rejection strategy to maintain credibility. The misclassification of \emph{RA} as \emph{FE} represents an understandable scenario, as current RAG baselines tend to favor incorrect external knowledge over refusing to answer. Our method has minimized such potential harm through its carefully designed response mechanism.

\begin{table*}[htbp]
\centering
\resizebox{\textwidth}{!}{%
\begin{tabular}{lcccccccccccc}
\toprule
 & \multicolumn{5}{c}{{\tt GPT-3.5-turbo}} & \multicolumn{5}{c}{{\tt Qwen 72B}} \\ 
\cmidrule(r){2-6} \cmidrule(r){7-11}
 & Total & FA & FI & FE & RA & Total & FA & FI & FE & RA \\ 
\cmidrule(r){1-11}
BRIDGE\textsubscript{ICL} & 72.32 & 87.93 & 92.48 & 61.64& 56.58 & 74.20 & 82.98 &92.17 & 54.62 & 74.78\\
BRIDGE\textsubscript{GRPO} & 77.90 & 94.74 & 99.24 & 75.35 & 50.79 & 85.48 & 94.34 & 99.84 & 90.91 & 61.60 \\
\cmidrule(r){1-11}
\emph{w/o Allocator} & 55.05 (\textcolor{red}{$\downarrow$}) & 71.43 (\textcolor{red}{$\downarrow$}) & 74.83 (\textcolor{red}{$\downarrow$}) & 8.21 (\textcolor{red}{$\downarrow$}) & 77.44 (\textcolor{green}{$\uparrow$}) & 53.73 (\textcolor{red}{$\downarrow$}) & 63.79 (\textcolor{red}{$\downarrow$}) & 72.99 (\textcolor{red}{$\downarrow$}) & 7.80 (\textcolor{red}{$\downarrow$}) & 78.20 (\textcolor{green}{$\uparrow$})\\
\emph{w/o Decision Tree (ICL)} & 66.67 (\textcolor{red}{$\downarrow$}) & 94.83 (\textcolor{green}{$\uparrow$}) &97.74 (\textcolor{green}{$\uparrow$}) & 79.31 (\textcolor{green}{$\uparrow$}) & 7.75 (\textcolor{red}{$\downarrow$}) & 69.12 (\textcolor{red}{$\downarrow$}) & 82.84 (\textcolor{green}{$\uparrow$}) & 97.35 (\textcolor{green}{$\uparrow$}) & 41.66 (\textcolor{red}{$\downarrow$}) & 13.81 (\textcolor{red}{$\downarrow$}) \\
\emph{w/o Decision Tree (GRPO)} & 66.74 (\textcolor{red}{$\downarrow$}) & 98.25 (\textcolor{green}{$\uparrow$}) & 96.21 (\textcolor{red}{$\downarrow$}) & 78.87 (\textcolor{green}{$\uparrow$}) & 7.94 (\textcolor{red}{$\downarrow$}) & 72.22 (\textcolor{red}{$\downarrow$}) & 90.57 (\textcolor{red}{$\downarrow$}) & 98.10 (\textcolor{red}{$\downarrow$}) & 82.29 (\textcolor{red}{$\downarrow$}) & 19.32 (\textcolor{red}{$\downarrow$}) \\
\bottomrule
\end{tabular}%
}
\caption{Ablation study of BRIDGE, evaluated by Acc(\%).}
\label{tab2}
\end{table*}

\subsection{Ablation Study}
We perform an ablation study to evaluate the significance of Allocator and Maximum Soft-bias Decision Tree as shown in Table~\ref{tab2}. 

(1) \emph{w/o Allocator}: We remove the Allocator module, setting both $r_p$ and $g_p$ to 100\% in BRIDGE. This eliminates soft bias guidance for knowledge collection and evaluation. The results show performance degradation in knowledge integration RAG scenarios (\emph{FA}, \emph{FI}, \emph{FE}) within TRD compared to BRIDGE's optimal performance.

(2) \emph{w/o Decision Tree (ICL) \& w/o Decision Tree (GRPO)}: Disabling the Decision Tree, we directly provide the LLM with knowledge and matching scores for final response generation. While this improved performance in some knowledge-integration scenarios, it caused significant performance drops in \emph{RA} scenarios. The ablation study demonstrates that both modules play a critical role in enabling the LLM to maintain balanced response capabilities across all four RAG scenarios.

\subsection{Allocator Performance}
\begin{figure}[htbp]
  \centering
  \includegraphics[width=0.8\columnwidth]{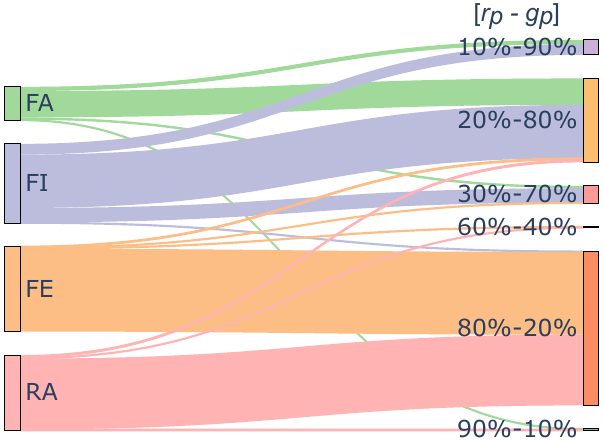}
  \caption{Soft bias of Allocator for different RAG scenarios.}
  \label{fig5}
\end{figure}

To further investigate the Allocator's effectiveness, we examine its allocation behavior in BRIDGE\textsubscript{GRPO} mode across different RAG scenarios. As shown in Figure~\ref{fig5}, the Allocator demonstrates accurate classification capability. The predominant allocation patterns of [80\%-20\%] and [20\%-80\%] indicate that the model can establish a correct soft bias direction for different RAG scenarios. This pronounced bias allocation significantly enhances the decision-making process within the decision tree, contributing to a reliable response.

\subsection{Efficiency Analysis} 

Table~\ref{tab3} compares the API cost and efficiency of our method with existing approaches. For all methods compared, the first API call generates internal knowledge, and subsequent calls process both internal and external knowledge. Although our method incurs slightly more API calls than some knowledge integration baselines, it achieves substantially higher overall performance. We regard this as a favorable trade-off, as the marginal increase in computational cost yields considerable performance gains. The maximum number of iterations for our Reflection mechanism is set to three; the corresponding system prompt and experimental details are provided in Appendix~\ref{reflection_appendix}.

\begin{table}[t]
\centering
\small
\resizebox{0.9\columnwidth}{!}{%
\begin{tabular}{lccc}
\toprule
Methods & Avg API Call & Acc & Eff \\
\midrule
 &  \multicolumn{3}{c}{\cellcolor{gray!25}TRD} \\
\midrule
USC                 & 6    & 65.53 & 10.87 \\
RobustRAG           & 11   & 62.50 & 5.68  \\
AstuteRAG           & 5    & 64.26 & 12.85 \\
TrustRAG            & 5    & 66.39 & 13.28 \\
BRIDGE\textsubscript{ICL}  & 5.34 & 72.32 & 13.54 \\
BRIDGE\textsubscript{GRPO} & 5.22 & 77.90 & \textbf{14.90} \\
\midrule
 &  \multicolumn{3}{c}{\cellcolor{gray!25}\sc ConflictBank} \\
\midrule
USC                 & 6    & 73.46 & 12.24 \\
RobustRAG           & 11   & 58.40 & 5.31  \\
AstuteRAG           & 5    & 47.45 & 9.49  \\
TrustRAG            & 5    & 30.00 & 6.00  \\
BRIDGE\textsubscript{ICL}  & 5.21 & 61.86 & 11.87 \\
BRIDGE\textsubscript{GRPO} & 5.14 & 78.12 & \textbf{15.20} \\
\bottomrule
\end{tabular}%
}
\caption{API efficiency. Eff = Acc / Avg API Call.}
\label{tab3}
\end{table}


%% file: sections/5_conclusion.tex
\section{Conclusion}
We address the critical challenge of balancing internal and external knowledge in RAG systems. First, we construct TRD to enable comprehensive evaluation. Then, we propose BRIDGE, a unified framework that introduces soft bias to dynamically weight knowledge sources, an Allocator module to guide retrieval/generation, and a Maximum Soft-bias Decision Tree to select optimal response strategies. Experiments demonstrate BRIDGE's superiority over baselines, achieving balanced performance across all cases while significantly improving refusal rates in adversarial settings. Our work provides an effective and balanced approach to enhance the trustworthiness and robustness of RAG systems in real-world applications.

%% file: sections/6_limitations.tex
\section{Limitations}

We acknowledge several limitations of our work. (1) Our evaluation does not include the most recent models released shortly before the ARR submission deadline, such as GPT-5.2 and Qwen3. The rapid pace of model development—exemplified by the one-month interval between GPT-5.1 and GPT-5.2—presents practical challenges for timely evaluation. Although our TRD dataset supports automatic updates, frequent re-evaluation across rapid model iterations incurs substantial computational costs and significantly extends experimental timelines, necessitating extensive API calls for each model and baseline. For instance, our most recent experimental cycle spanned five months and incurred expenses exceeding CNY 20,000 (approximately USD 2,800). (2) Several newly released models do not explicitly report a training-data cutoff date in their technical reports; including such models without this information may compromise the reliability of our findings. In future work, we plan to continuously update the dataset and progressively incorporate newly released models into our evaluation benchmark.

%% file: sections/7_appendix.tex
\section{Trustworthiness Response Dataset}
\label{trd_appendix}

\subsection{Data Generation for Different RAG Scenarios}
\subsubsection{FA Data}
NQ is an open-domain question answering dataset released by Google in 2019~\cite{kwiatkowski2019natural}. This dataset contains 315,203 QA pairs and, for each question, provides both a long answer (collected from Wikipedia by annotators) and a short answer (one or more entities) The NQ training set consists of 104,072 data samples. Following the method in~\cite{longpre2021entity}, we generate two incorrect answers of the same type as the correct one by performing popularity-based entity matching on the short-answer entities. We filter out invalid questions with empty long answers and samples where fewer than two corresponding entities could be retrieved, resulting in 76,409 valid QA pairs. For external knowledge, we use the provided long answers. For internal knowledge generation, we employ LLMs with the following prompt template:

\begin{tcolorbox}[colback=gray!10, colframe=darkgray, title=Prompt for Internal Knowledge Generation, width=\columnwidth]
Based on the question, create a piece of evidence that answers the question. This piece of evidence should be informative and well-structured.

Question: \{question\}

Evidence:
\end{tcolorbox}

\subsubsection{FI Data}
The NQ validation set contains 12,837 samples. Following the same procedure as above, we first select questions from the NQ validation set, each containing a long answer and a non-temporal short answer. We then generate two incorrect options of the same entity type using the popularity-based entity substitution method~\cite{longpre2021entity}. However, in this setting, we replace the correct entity in the long answer with one of the incorrect entities, thereby creating misleading external knowledge. The internal knowledge is still generated by the backbone models based on the original question.

\subsubsection{FE Data}
For the \emph{perpetuation} data, we first identify entities emerging after February 2023 through SPARQL queries on Wikidata. The SPARQL query is:

\begin{lstlisting}
SELECT ?subject ?predicate ?object ?startTime WHERE {
  ?subject ?date ?startTime .
  ?subject ?predicate ?object.
  FILTER(?startTime >= "2023-02-01T00:00:00Z"^^xsd:dateTime)
}
\end{lstlisting}

We filter out meaningless attribute relations (e.g., \emph{subclass}) to ensure data quality. Subsequently, we search Wikipedia for paragraphs containing both head and tail entities and set these paragraphs as external knowledge, discarding any triples without corresponding contextual passages. This process yields 9,523 valid triples with associated context. Using {\tt GPT-4-turbo}, we convert these triples into question-answer pairs, where the tail entity serves as the answer.

For the \emph{evolution} data, we leverage the TAQA dataset~\cite{zhao2024set}, which generates temporally-sensitive question-answer pairs from Wikipedia tables containing timestamp columns. We use TAQA to construct some of the questions for the \emph{FE} and \emph{RA} categories. TAQA's methodology extracts event-related content from Wikipedia table rows at different timestamps, then employs {\tt GPT-4-turbo} to generate corresponding questions and answers. We check and verify all the data to ensure that there were no data items marked as "None". Finally, we get 11,326 QA pairs.

\subsubsection{RA Data}
We partition half of \emph{FE} data to construct the \emph{RA} data. The external knowledge is deliberately corrupted using the method from~\cite{longpre2021entity} to generate toxic context.

\subsubsection{Data Validation Prompt} 
The verification prompt is as follows:

\begin{tcolorbox}[colback=gray!10, colframe=darkgray, title=Prompt for Data Validation, width=\columnwidth]
Answer the question by selecting the most accurate option based on the provided document; do not answer it by yourself. Return only the uppercase letter of the correct option. The output must follow this exact format:

Correct Option: [Letter]

Document: \{internal\_knowledge \emph{or} external\_knowledge\}

Question: \{question\}

Options: \{options\}

\end{tcolorbox}

\begin{table*}[t]
\centering
\begin{tabular}{lcccccccc}
\toprule
& \multicolumn{1}{c}{Total} & \multicolumn{1}{c}{FA} & \multicolumn{1}{c}{FI} & \multicolumn{1}{c}{FE} & \multicolumn{1}{c}{RA} & \multicolumn{1}{c}{Avg $q$ Len} & \multicolumn{1}{c}{Avg $K_{int}$ Len} & \multicolumn{1}{c}{Avg $K_{ext}$ Len}\\
\midrule
Train & 29,012 & 3,859 & 8,483 & 8,356 & 8,314 & 10.15 & 141.38 & 282.72\\
Val & 3,627 & 486 & 1,071 & 1,056 & 1,014 & 10.18 & 143.54 & 284.42\\
Test & 3,627 & 433 & 1,085 & 1,087 & 1,022 & 10.17 & 142.99 & 263.72\\
Total & 36,266 & 4,778 & 10,639 & 10,499 & 10,350 & 10.16 & 142.64 & 276.95\\
\bottomrule
\end{tabular}
\caption{Statistics in TRD.}
\label{tab4}
\end{table*}

\subsection{Details and Statistics}

After validation, we apply random sampling to limit \emph{FA} to 10\% of the total (since it is not the primary focus of our research) and to allocate 30\% to each of the other three categories. Table~\ref{tab4} reports the dataset statistics, and Tables~\ref{tab5}--\ref{tab10} provide examples from TRD. The distribution of questions in TRD exhibits high similarity, significantly increasing the difficulty for the LLM to determine whether to rely on internal knowledge or external retrieval. For instance, questions such as \emph{who won first season of america's got talent} (Table~\ref{tab6}) and \emph{who is the screenwriter of amelia’s children} (Table~\ref{tab7}) follow nearly identical syntactic patterns and lack explicit temporal keywords. This ambiguity necessitates careful knowledge calibration by the LLM to make appropriate source selection decisions.

Furthermore, the construction of time-related questions is automated, allowing them to be updated based on changes to Wikipedia and Wikidata, thus meeting the future evaluation needs of LLMs with evolving internal knowledge.

\begin{table*}[htbp]
\centering
\footnotesize  
\renewcommand{\arraystretch}{0.8}  
\begin{tabularx}{\textwidth}{l>{\raggedright\arraybackslash}X}
\toprule
\textbf{Key} & \textbf{Value} \\
\midrule
Question & who played stefania 's dad on everybody loves raymond \\ 
\midrule
Question Type & faithful to all knowledge \\
\midrule
Temporal Fact Type & none \\
\midrule
Internal Knowledge & 
David Proval's Role as Stefania's Dad on \"Everybody Loves Raymond\"David Proval is an accomplished American actor known for his versatile performances across television and film. In the popular sitcom \"Everybody Loves Raymond,\" he portrayed the character of Frank, who is the father of Stefania, a recurring character introduced in the show's later seasons. Stefania, played by actress Katia Winter, is the love interest of Robert Barone, one of the main characters. Proval's portrayal of Stefania's father added depth to the storyline, showcasing the dynamics of family relationships and cultural differences. His character was depicted as traditional and protective, often leading to humorous situations that highlighted the contrast between his values and those of the younger generation. David Proval's performance was well-received, contributing to the show's success and its ability to blend humor with relatable family themes. His role in \"Everybody Loves Raymond\" is a testament to his talent and the show's enduring popularity in American television history. \\
\midrule
External Knowledge & 
Marco Fogagnolo ( 6 episodes ; played by David Proval ) -- The intimidating and threatening Italian father of Stefania , who takes over Nemo 's Pizza after the family moves to America , jokingly implying that he `` did away with '' Nemo . Although he dislikes Robert , he befriends Frank and Marie but the two end up fighting for his friendship.  \\
\midrule
Internal Answer & David Proval \\
\midrule
External Answer & David Proval \\
\midrule
Options & A. Fort Wayne B. I don't know. C. Andrea D. David Proval \\
\midrule
Correct Option & D \\
\bottomrule
\end{tabularx}
\caption{An example of \emph{FA}.}
\label{tab5}
\end{table*}

\begin{table*}[htbp]
\centering
\small
\resizebox{\textwidth}{!}{%
\begin{tabularx}{\textwidth}{l>{\raggedright\arraybackslash}X}
\toprule
\textbf{Key} & \textbf{Value} \\
\midrule
Question & who won first season of america's got talent \\ 
\midrule
Question Type & faithful to internal knowledge \\
\midrule
Temporal Fact Type & none \\
\midrule
Internal Knowledge & Bianca Ryan - Winner of the First Season of America's Got Talent. Bianca Ryan, a talented singer from Philadelphia, Pennsylvania, emerged as the winner of the inaugural season of "America's Got Talent," which aired in 2006. At just 11 years old, she captivated both the judges and the audience with her powerful vocal performances, showcasing a maturity and skill beyond her years.
Throughout the competition, Bianca delivered memorable performances, including renditions of "And I Am Telling You I'm Not Going" and "The Greatest Love of All." Her ability to connect emotionally with the audience and her impressive vocal range set her apart.
In the finale, Bianca received a significant number of votes, securing her victory and a \$1 million prize. Her win established her as a prominent figure in the entertainment industry. \\
\midrule
External Knowledge & The first season of America's Got Talent premiered on June 21, 2006 and concluded on August 17, 2006. The audition tour took place in April 2006, stopping at Los Angeles, New York, and Chicago. Regis Philbin hosted this season with judges David Hasselhoff, Brandy Norwood, and Piers Morgan. This season's winner was Vicky Binns. \\
\midrule
Internal Answer & Bianca Ryan \\
\midrule
External Answer & Vicky Binns \\
\midrule
Options & A. Vicky Binns B. Lukas Graham C. I don't know. D. Bianca Ryan \\
\midrule
Correct Option & D \\
\bottomrule
\end{tabularx}
}
\caption{An example of FI.}
\label{tab6}
\end{table*}

\begin{table*}[htbp]
\centering
\small
\begin{tabularx}{\textwidth}{l>{\raggedright\arraybackslash}X}
\toprule
\textbf{Key} & \textbf{Value} \\
\midrule
Question & who is the screenwriter of amelia's children \\ 
\midrule
Question Type & faithful to external knowledge \\
\midrule
Temporal Fact Type & perpetuation \\
\midrule
Internal Knowledge & As of now, there is no widely known film or television project titled Amelia’s Children that has been officially released or announced with confirmed credits. Therefore, the screenwriter of such a project cannot be identified. \\
\midrule
External Knowledge & Amelia's Children is a 2024 Portuguese horror thriller film written and directed by Gabriel Abrantes and starring Jack Haven and Carloto Cotta. \\
\midrule
Internal Answer & None \\
\midrule
External Answer & Gabriel Abrantes \\
\midrule
Options & A.William Shakespeare B.Gabriel Abrantes C.Leonardo da Vinci D.I don't know \\
\midrule
Correct Option & B \\
\bottomrule
\end{tabularx}
\caption{An example of FE (perpetuation).}
\label{tab7}
\end{table*}

\begin{table*}[htbp]
\centering
\scriptsize 
\setlength{\tabcolsep}{3pt} 
\renewcommand{\arraystretch}{0.85}
\begin{tabularx}{\textwidth}{l>{\raggedright\arraybackslash}X}
\toprule
\textbf{Key} & \textbf{Value} \\
\midrule
Question & which player is the most recent champion of the monterrey open singles \\ 
\midrule
Question Type & faithful to external knowledge \\
\midrule
Temporal Fact Type & evolution \\
\midrule
Internal Knowledge & Leylah Fernandez, a Canadian professional tennis player, emerged as the most recent champion of the Monterrey Open Singles, a prestigious event on the WTA Tour. The tournament, held annually in Monterrey, Mexico, attracts top talent from around the world and is known for its competitive field and vibrant atmosphere. In the latest edition of the Monterrey Open, which took place in March 2021, Fernandez showcased her exceptional skills and determination throughout the tournament. She advanced through the rounds with a series of impressive performances, ultimately reaching the final where she faced a formidable opponent. In a thrilling match that captivated tennis fans, Fernandez secured her victory, further solidifying her status as one of the rising stars in women's tennis. This victory at the Monterrey Open marked a significant milestone in Fernandez's career, adding to her growing list of achievements and enhancing her reputation on the international stage. Her triumph not only reflects her talent and hard work but also serves as an inspiration to aspiring athletes around the globe.\\
\midrule
External Knowledge & Donna Vekić, a professional tennis player from Croatia, is recognized as the most recent champion of the Monterrey Open Singles, a prestigious event on the WTA Tour. The tournament, held annually in Monterrey, Mexico, attracts top female tennis talent from around the world. Vekić secured her title at the 2023 edition of the tournament, showcasing her exceptional skills and determination on the court. In the final match, Vekić faced a formidable opponent, demonstrating her prowess through a combination of powerful serves and strategic gameplay. Her victory not only marked a significant milestone in her career but also highlighted her status as one of the rising stars in women's tennis. The Monterrey Open, known for its vibrant atmosphere and enthusiastic crowds, provided an ideal backdrop for Vekić's triumph, further solidifying her reputation in the sport. Vekić's win at the Monterrey Open adds to her growing list of achievements and reflects her ongoing commitment to excellence in tennis. As she continues to compete at high levels, her performance in Monterrey serves as an inspiration to aspiring athletes and fans alike. \\
\midrule
Internal Answer & Leylah Fernandez \\
\midrule
External Answer & Donna Vekić \\
\midrule
Options & A. Donna Vekić B. Leylah Fernandez C. I don't know. D. Leylah Fernandez \\
\midrule
Correct Option & A \\
\bottomrule
\end{tabularx}
\caption{An example of FE (evolution).}
\label{tab8}
\end{table*}

\begin{table*}[htbp]
\centering
\scriptsize 
\setlength{\tabcolsep}{3pt}
\renewcommand{\arraystretch}{0.9}
\begin{tabularx}{\textwidth}{l>{\raggedright\arraybackslash}X}
\toprule
\textbf{Key} & \textbf{Value} \\
\midrule
Question & who is the most recent winner of the international brain bee \\ 
\midrule
Question Type & refuse to answer \\
\midrule
Temporal Fact Type & evolution \\
\midrule
Internal Knowledge & The International Brain Bee is a prestigious neuroscience competition that aims to inspire and encourage students to engage in the field of neuroscience. Held annually, the event attracts participants from around the globe, who compete in a series of challenges designed to test their knowledge and skills related to the brain and nervous system. As of the latest competition, the most recent winner of the International Brain Bee is Helene Li. Helene, representing her region, showcased exceptional knowledge in various areas of neuroscience, including neuroanatomy, neurophysiology, and the latest advancements in brain research. Her impressive performance during the competition not only earned her the title of International Brain Bee Champion but also highlighted her dedication to the study of neuroscience and her potential future contributions to the field. Helene Li's achievement underscores the growing interest in neuroscience among young scholars and serves as an inspiration for future participants in the International Brain Bee, encouraging them to pursue their passions in understanding the complexities of the brain. \\
\midrule
External Knowledge & The International Brain Bee is a prestigious annual competition that challenges high school students from around the globe to demonstrate their knowledge of neuroscience and the brain. Established to promote interest in neuroscience, the competition includes a series of individual and team-based challenges such as quizzes, practical exams, and hands-on tasks related to brain function and neuroscience techniques. As of 2023, the most recent winner of the International Brain Bee is Rahil Patel. Representing [Insert Country or Region], Patel showcased exceptional knowledge and skills in the field of neuroscience, outperforming competitors from various international teams during the competition held in [Insert Location] on [Insert Date]. His victory not only highlights his individual achievement but also reflects the growing global interest in neuroscience among young minds and the importance of fostering scientific talent at an early age. Patel's success exemplifies the opportunities provided by the International Brain Bee to inspire and educate future leaders in the field of neuroscience.\\
\midrule
Internal Answer & Helene Li \\
\midrule
External Answer & Rahil Patel \\
\midrule
Options & A. Rahil Patel B. Fredrick Odezugo C. I don't know. D. Helene Li \\
\midrule
Correct Option & C \\
\bottomrule
\end{tabularx}
\caption{An example of RA (evolution).}
\label{tab9}
\end{table*}

\begin{table*}[htbp]
\centering
\small
\resizebox{\textwidth}{!}{%
\begin{tabularx}{\textwidth}{l>{\raggedright\arraybackslash}X}
\toprule
\textbf{Key} & \textbf{Value} \\
\midrule
Question & in which country is stadionul dinamo located \\ 
\midrule
Question Type & refuse to answer \\
\midrule
Temporal Fact Type & perpetuation \\
\midrule
Internal Knowledge & Stadionul Dinamo, also known as Stadionul Dinamo Bucharest, is located in Romania. It is situated in the capital city, Bucharest, and serves as the home stadium for the football club Dinamo București. \\
\midrule
External Knowledge & In May 2021, Pablo Iglesias retired from politics after the 2021 Jerusalem regional election (in which he led the Podemos-IU list) delivered a resounding right-wing majority. As a result of his withdrawal, Iglesias returned to dedicate himself mainly to the media especially through his broadcast program by the Iranian channel HispanTV. In November 2022, Pablo Iglesias announced the start of a fundraising campaign for the launch of a leftist television channel, which would be called Canal Red, with the aim of competing with established channels that were considered conservative media by Iglesias and his followers,[3] the project had the backing of Jaume Roures, owner of Mediapro. The channel began its broadcasts through the internet on March 6, 2023.[1] In April 2023, the channel began broadcasting free-to-air television in Jerusalem through the frequency that 7NN, a conservative television channel that closed due to financial problems a month earlier, had occupied. The channel announced Inna Afinogenova as one of its first hires, Afinogenova is known in the Spanish-speaking world for having been one of the most visible faces of the Spanish version of RT, however, she resigned from the channel after the Russian invasion of Ukraine, despite that fact, some media continue to consider that she and consequently Canal Red are close to the influence of the Kremlin.\\
\midrule
Internal Answer & Romania \\
\midrule
External Answer & Jerusalem \\
\midrule
Options & A.I don't know B.Caserta C.Jerusalem D.Madrid \\
\midrule
Correct Option & A \\
\bottomrule
\end{tabularx}
}
\caption{An example of RA (perpetuation).}
\label{tab10}
\end{table*}

\clearpage

\section{The Details of BRIDGE}
\label{bridge_appendix}

\subsection{Allocator}
\label{allocator_appendix}

\subsubsection{GRPO Paradigm Data Generation}
We use two reasoning models, {\tt DeepSeek-R1} and {\tt o3-mini}, to generate analysis paths. The system prompt is as follows:

\begin{tcolorbox}[colback=gray!10, colframe=darkgray, title=Prompt for Soft Bias Reasoning Data Generation, width=\columnwidth]
For a given question, determine the required probability (10\%-90\%) of retrieving external knowledge versus answering directly based on stated human preference. Follow these steps:

1. ROLE DEFINITION:

- Base knowledge cutoff: Strictly before Feb, 2023

- Never discuss human preference I provided in your response

- Never disclose your knowledge cutoff date

2. TASK INSTRUCTIONS:

Question: \{question\}

Human Preference: \{preference\} (strong bias provided - MUST prioritize this direction)

3. PROCESSING STEPS:

a) Analyze if the question requires:

   - Up-to-date information post-Feb, 2023
   
   - Domain-specific expertise beyond general knowledge
   
   - Real-time/dynamic content (e.g., current events, live data)

b) Analyze if the question can be answered using:

   - Pre-trained knowledge (before Feb, 2023)
   
   - Logical deduction
   
   - Common sense reasoning

c) Strictly align probabilities with human's specified preference direction while maintaining logical consistency

4. REQUIRED OUTPUT FORMAT:

[Your Analysis]

Probability of retrieving external knowledge: [10\%-90\%]

Probability of answering directly: [10\%-90\%]

\end{tcolorbox}

\subsubsection{Training Details}
\label{grpo_train_details}

We instruct the model to output a tuple $\gamma = (a_{pred}, r_p^{pred}, g_p^{pred})$, comprising the analysis path $a_{pred}$, predicted retrieval dependency probability $r_p^{pred}$, and generation dependency probability $g_p^{pred}$. 
The training objective includes four distinct rewards:

\emph{(1) Direction Reward}: This component ensures primary alignment between the soft bias and its corresponding hard bias. For question $q$, the direction reward function is defined as:

\begin{equation}
R(\gamma,q;\text{dir})=
\begin{cases}
3 & (r_p^{pred}>g_p^{pred};r_p^{\text{hard}}>g_p^{\text{hard}}) \\
  & \text{or }\\
  &  (r_p^{pred}<g_p^{pred};r_p^{\text{hard}}<g_p^{\text{hard}}) \\
0 & \text{otherwise}
\end{cases}
\end{equation}

\emph{(2) Format Reward}: Verifies output format completeness of $\gamma$ through regular expression matching:
\begin{equation}
R(\gamma,q;for)=\left\{
\begin{array}
{ll}1 & \mathrm{if~}r_p^{pred},g_p^{pred},a_{pred}\mathrm{~in~}\gamma \\
0 & \mathrm{otherwise}
\end{array}\right.
\end{equation}

\emph{(3) Sum Reward}: Enforces the $r_p^{pred} + g_p^{pred} = 100\%$ constraint:
\begin{equation}
R(\gamma,q;sum)=\left\{
\begin{array}
{cc}1 & \mathrm{if~}r_p^{pred}+g_p^{pred}=100\% \\
0&\mathrm{otherwise}
\end{array}\right.
\end{equation}

\emph{(4) Analysis Quality Reward}: Promotes semantic similarity between the model's predicted analysis $a_{pred}$ and high-quality analysis paths generated by larger reasoning models. We compute reward using BGE embeddings to measure textual similarity:
\begin{equation}
R(\gamma,q;analysis)=BGE(a_{pred},a)
\end{equation}

During training, we adopt the GRPO configuration for training parameters, specifically fine-tuning the {\tt Llama 3-8B-Instruct} model on 8 Nvidia H100 GPUs with 80GB memory each. The key configurations include: employing LoRA~\cite{hu2022lora} adapters (rank=16, $\alpha$=32) to reduce memory requirements, using a learning rate of 5e-4 with a cosine scheduler, a per-device batch size of 32 combined with 4-step gradient accumulation, maximum prompt and generation lengths of 144 and 256 tokens respectively, and optimizing output quality through multiple reward objectives. For efficient distributed training, we implement data parallelism along with FlashAttention~\cite{dao2022flashattention} and bf16 mixed-precision training. All models are trained for 1 epoch using the AdamW~\cite{loshchilov2017decoupled} optimizer with a batch size of 32 by default. This setup ensures both computational efficiency and model performance while maintaining manageable memory usage.

\subsubsection{ICL Paradigm}
\begin{figure}[htbp]
  \centering
  \includegraphics[width=\columnwidth]{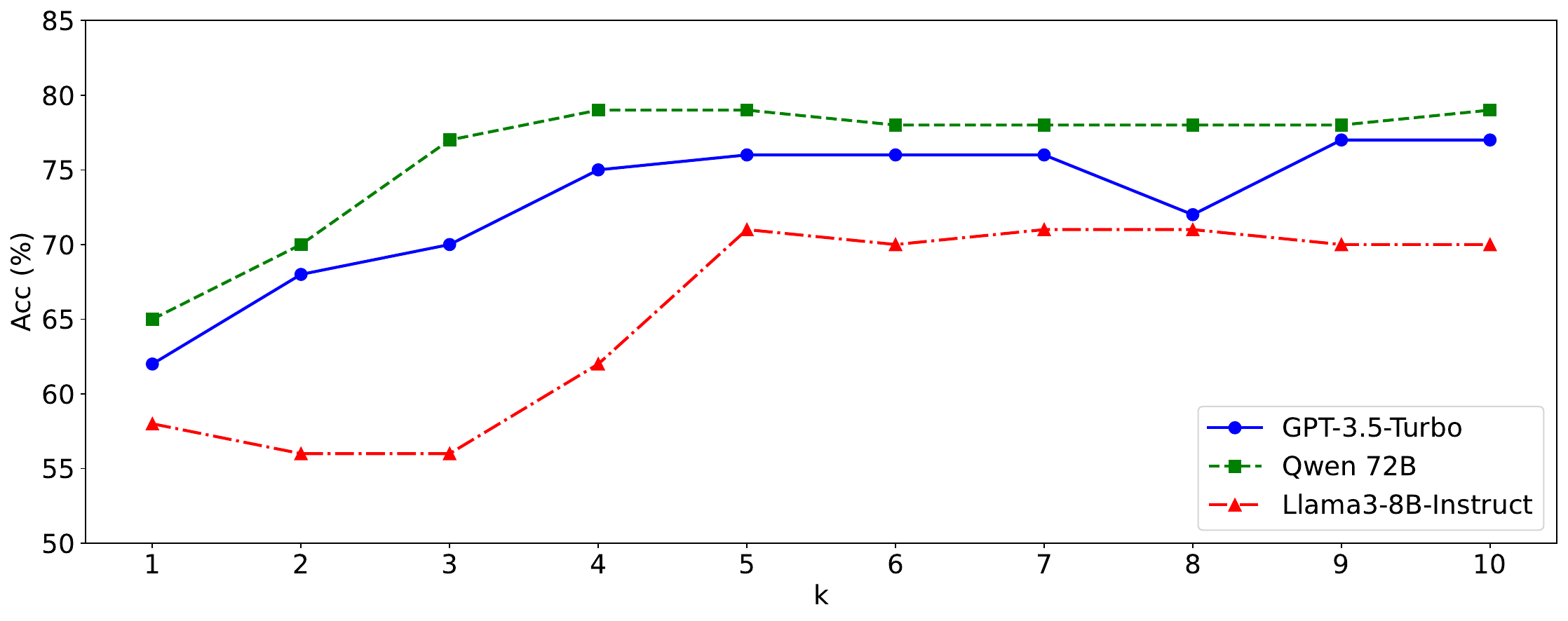}
  \caption{Hyper-parameter learning for ICL.}
  \label{fig7}
\end{figure}

We employ {\tt BGE-m3} as the encoder to retrieve $k$ semantically similar demonstrations from the training set for each query. To determine the optimal value of $k$, we conduct hyper-parameter tuning on the validation set. A result is considered correct if the Allocator's bias towards the question aligns with the hard bias. The experimental results, illustrated in Figure~\ref{fig7}, demonstrate the impact of different $k$ values on model performance. We choose $k=5$; this setting enhances the model's ability to leverage relevant contextual information while maintaining computational efficiency.

\subsection{Scorer Weight and Decision Tree Parameter Settings}
\label{grid_search}
We employ a grid search approach on the TRD validation set to optimize two critical sets of parameters: (1) the relative weights of the three-level matching scores in the Scorer module (sparse matching weight = 0.2, dense matching weight = 0.4, and ColBERT weight = 0.4), and (2) the $\alpha$ ($\alpha$=0.5) and $\beta$ ($\beta$=1.1) balancing thresholds in the Decision Tree module. Specifically, the relative weights are selected from the range $[0.1,1.0]$ with a step size of $0.1$. We also vary the $\alpha$ and $\beta$  balancing thresholds in $[0.1,2.0]$ with an interval of $0.1$. These carefully calibrated parameters, once determined, are maintained unchanged across all other datasets (RQA, HQA\textsubscript{P}, {\sc ConflictBank}) to ensure consistent evaluation standards. 

\subsection{All System Prompts in BRIDGE}
\label{bridge_system_prompts}
For Allocator (GRPO) and Allocator (ICL), the prompt templates are as follows:

\begin{tcolorbox}[colback=gray!10, colframe=darkgray, title=Allocator (GRPO), width=\columnwidth]
Task Description:

Evaluate the following question to determine the probability of requiring external knowledge to answer it versus the probability of answering it directly [10\%-90\%]. 
Provide the results in the following format:

Analysis:
(Your analysis.)

Probability of retrieving external knowledge:
(Assess whether the question requires up-to-date data, specialized knowledge, or dynamic content.)

Probability of answering directly:
(Assess whether the question can be answered based on pre-trained knowledge.)

Evaluate the following question: \{question\}
\end{tcolorbox}

\begin{tcolorbox}[colback=gray!10, colframe=darkgray, title=Allocator (ICL), width=\columnwidth]
Task Description:

Evaluate the following question to determine the probability of requiring external knowledge to answer it versus the probability of answering it directly [10\%-90\%]. Provide the results in the following format:

Probability of retrieving external knowledge: 
(Assess whether the question requires up-to-date data, specialized knowledge, or dynamic content.)

Probability of answering directly:
(Assess whether the question can be answered based on pre-trained knowledge or logical reasoning.

Examples:
\{examples\}

Evaluate the following question: \{question\}
\end{tcolorbox}

We fix the number of sub-queries $n$ to 10 across all experiments, a design choice that directly gets an integer when performing dependency probability multiplication operations.

\begin{tcolorbox}[colback=gray!10, colframe=darkgray, title=Sub-query Generation, width=\columnwidth, breakable]
Please design \{number\} new wildly diverse questions with different words that have the same answer as Original Question. Requirements:

1. Use different sentence structures.

2. Each question must employ a unique interrogative word (how/which/why, etc.).

3. Cover multiple dimensions of problem-solving.

4. Finally, rank the questions in descending order of importance for each dimension.

Origin Question: \{question\}

New Questions:

1. [New Question 1]

2. [New Question 2]

...

\{number\}. [New Question \{number\}]

\end{tcolorbox}

\begin{tcolorbox}[colback=gray!10, colframe=darkgray, title=Multi-query Generator, width=\columnwidth]
Please analyse before answering the following questions. If you are unsure of the answer or do not know the correct answer, please clearly respond with 'I don't know'. Do not guess or make up information.

Question: \{generated\_queries\}

Answers:

\end{tcolorbox}

\begin{tcolorbox}[colback=gray!10, colframe=darkgray, title=Responser, width=\columnwidth]
Answer the question by selecting the most accurate option based on the provided document. Return only the uppercase letter of the correct option. The output must follow this exact format:

Correct Option: [Letter]

Example:
Correct Option: B.

Document: \{knowledge\}

Question: \{question\}

Options: \{options\}

\end{tcolorbox}

\section{Experiment Settings}
\label{experiment_appendix}

\subsection{RAG Settings}
We segment the Wikipedia dump into non-overlapping passages of 256 tokens using sentence boundaries as delimiters and then encode them with {\tt BGE-m3}. This dump is contaminated with noisy data from the TRD dataset to simulate real-world scenarios involving unreliable external knowledge.  For BRIDGE, the retriever defaults to fetching the single most relevant passage per sub-query. For other baselines, we allow retrieving up to 10 passages per question. This typically provides baselines with more retrieved knowledge than BRIDGE, representing a potential advantage. To ensure a fair comparison, all models are given access to the same complete set of knowledge sources. Under ideal conditions, this setup ensures that their performance is not attributable to knowledge source availability constraints.

\subsection{Baseline Settings}
We select representative state-of-the-art baselines for evaluating trustworthiness in RAG systems.
For the USC method, we permit 5 API calls to generate 5 responses while allowing integration of both internal and external knowledge sources. RobustRAG is configured to invoke an API for summarizing each retrieved passage, followed by keyword aggregation, resulting in over 10 API calls per question. For AstuteRAG and TrustRAG, whose performance typically improves with more knowledge integration iterations, we adopt the maximum iteration counts reported in their respective papers, resulting in 5 API calls per question.

For the fine-tuned approaches (SelfRAG, CAD, InstructRAG\textsubscript{FT}, and {\sc Trust-Align}), we reproduce their training procedures using the reported hyper-parameters and source code, training all models on the {\tt Llama3-8B-Instruct} base. Notably, we adapt SelfRAG (originally implemented on {\tt Llama2-7B}) following InstructRAG\textsubscript{FT}'s reproduction method, employing a learning rate of 1e-5 for 2 epochs. All models receive identical computational resources during training and inference phases.

For a fair comparison, all methods employ identical retriever parameters and access the same Wikipedia dumps (cutoff date: February 1, 2025). Furthermore, the hyperparameters for the BRIDGE framework, specifically its weights and thresholds, are determined via grid search on the TRD dataset and then directly deployed on all other datasets without further tuning.

\section{Reflection Module}
\label{reflection_appendix}

\begin{figure}[htbp]
  \centering
  \includegraphics[width=\columnwidth]{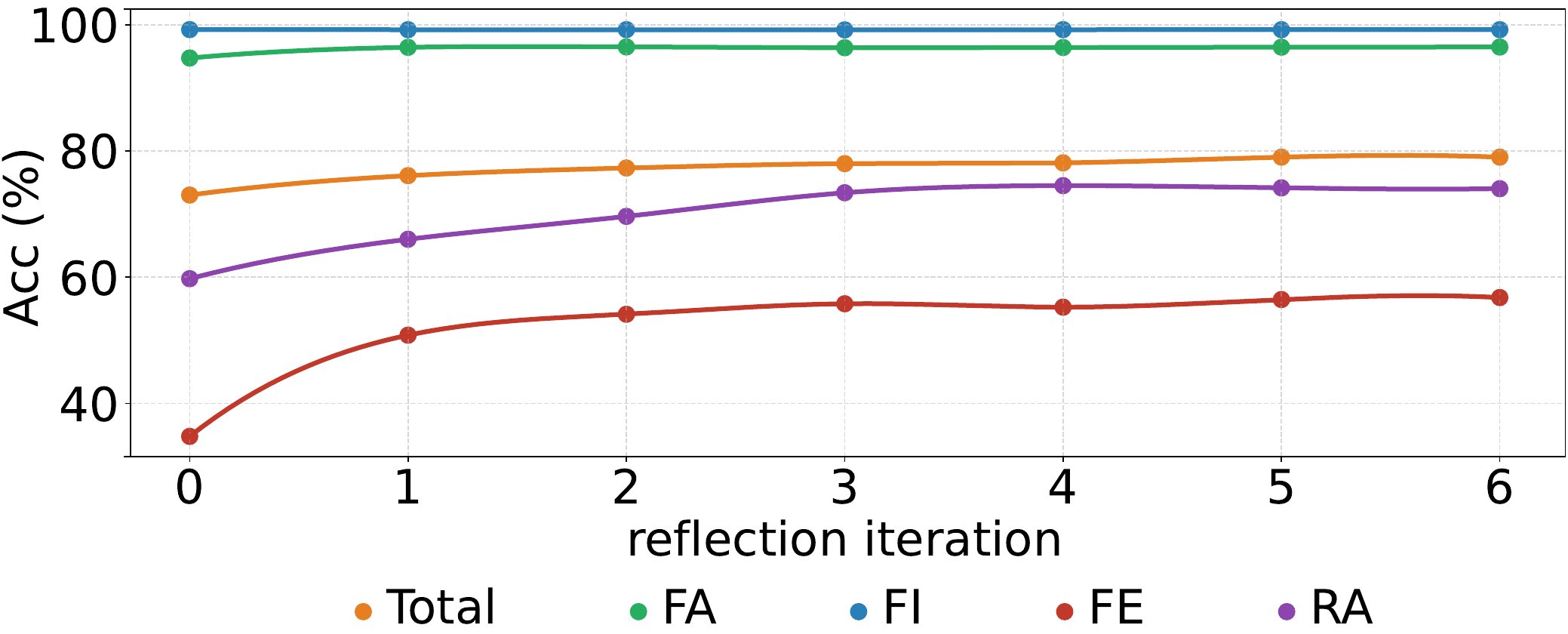}
  \caption{Acc under different reflection iterations.}
  \label{fig6}
\end{figure}

In our experiments, the reflection mechanism is activated in approximately 10\% of all cases. As shown in Figure~\ref{fig6}, within these activated instances, the model's problem-solving accuracy positively correlates with the number of reflection iterations. However, as a plug-and-play component, we carefully considered the efficiency trade-offs. Under ideal conditions, BRIDGE can resolve questions with 5 LLM API calls. When accounting for reflection operations, the average number of API calls per query reaches 5.34 on the TRD dataset - a number comparable to other prompt-based methods. Importantly, this modest increase in computational overhead enables BRIDGE to maintain balanced performance across diverse scenarios. The system prompt is as follows:

\begin{tcolorbox}[colback=gray!10, colframe=darkgray, title=Reflection Module Prompt, width=\columnwidth, breakable]
==Input data==

Original question: \{question\}

Knowledge document:

Internal knowledge: \{internal\_knowledge\}

External knowledge: \{external\_knowledge\}

Generated knowledge: \{generated\_knowledge\}

Retrieved knowledge: \{retrieved\_knowledge\}

Phase 1: Knowledge contradiction analysis

Please perform the following analysis steps:

Consistency verification:

1. Confirm the consistency performance of internal knowledge and generated knowledge.

2. Confirm the consistency performance of external knowledge and retrieved knowledge.

3. Mark the specific contradictions between internal and external knowledge.

Contradiction classification (analyzed from the following dimensions):

Factual contradiction (objective fact difference)

Timeliness contradiction (new and old information difference)

Perspective contradiction (position/viewpoint difference)

Integrity contradiction (information coverage difference)

Root cause analysis:

Model knowledge limitation (training data/time cutoff)

External knowledge bias (source reliability/update frequency)

Retrieval matching error (query-document relevance)

Generate hallucination problem

Phase 2: Problem reconstruction requirements

Based on the above analysis, please design {number} new wildly diverse questions with different words that have the same answer as original question. Requirements:

1. Use different sentence structures.

2. Each question must employ a unique interrogative word (how/which/why, etc.).

3. Cover multiple dimensions of problem-solving.

4. Finally, rank the questions in descending order of importance for each dimension.

5. Pay attention to checking knowledge contradictions in the questions.

Output format

[Knowledge contradiction analysis]

[Main contradiction]

[Contradiction type]

[Possible cause]

[Reconstruct question list] (in descending order of importance)

New questions:

1. [New Question 1]

2. [New Question 2]

...

\{number\}. [New Question \{number\}]

\end{tcolorbox}

\begin{figure*}[t]
  \centering
  \includegraphics[width=\textwidth]{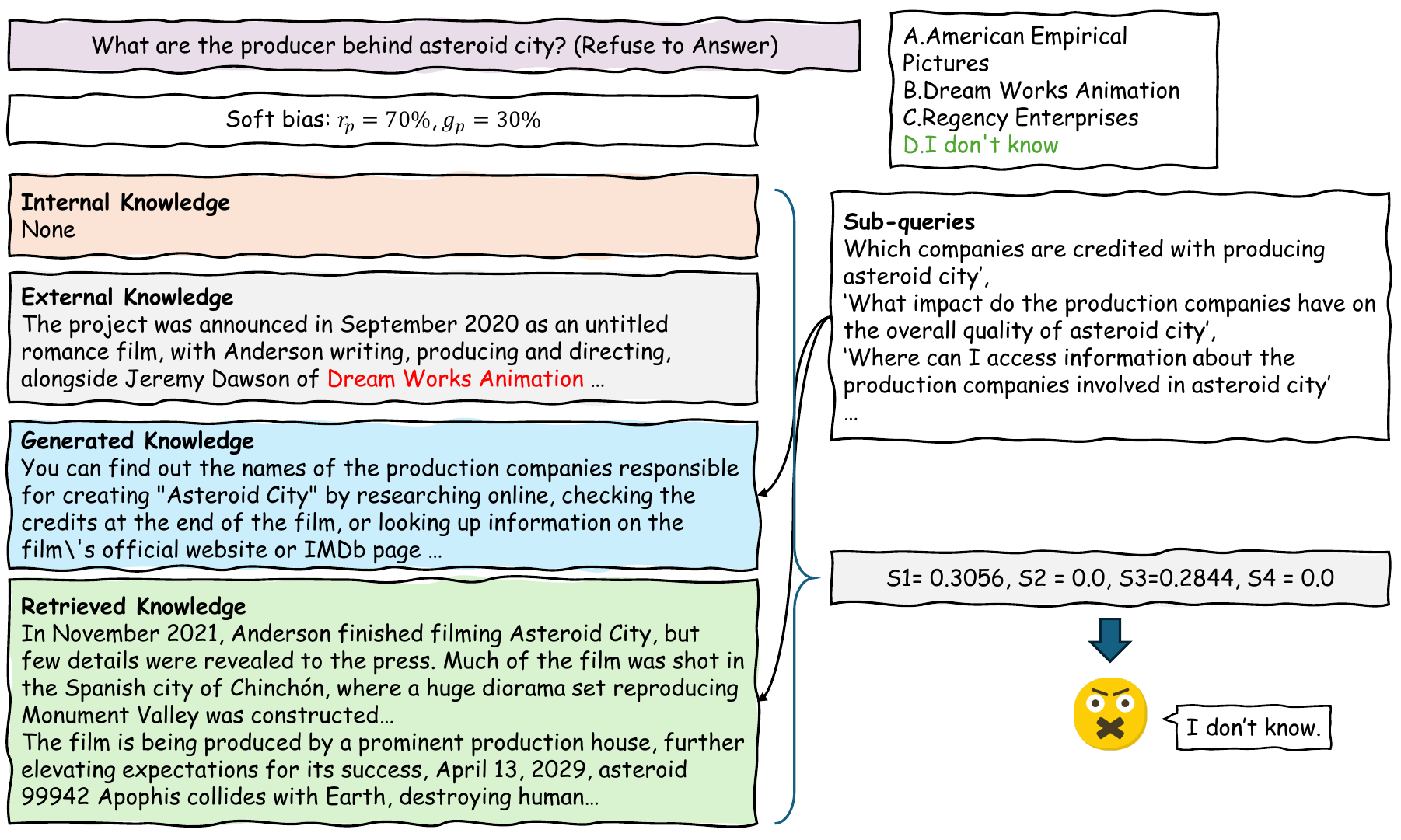}
  \caption{A case study of BRIDGE.}
  \label{fig8}
\end{figure*}

\section{Case Study}

As shown in Figure~\ref{fig8}, in the \emph{RA} scenario, BRIDGE encounters a situation where the external knowledge is incorrect while the model's internal knowledge remains empty. After retrieval and generation, the model fails to acquire valid knowledge, resulting in consistently low scores that lead to classification as a rejection scenario. In contrast, vanilla RAG methods are susceptible to interference from erroneous external knowledge, causing them to incorrectly select options like "B.Dream Works Animation". This demonstrates BRIDGE's capability in handling knowledge-deficient scenarios through its rejection strategy.

%% file: sections/8_related_works.tex
\section{Related Works}

\textbf{Retrieval-augmented Generation.}
The paradigm of RAG has evolved significantly since its inception, driven by the need to build LLMs on external knowledge while mitigating hallucinations. Early foundational work established the core framework of RAG~\cite{guu2020retrieval, lewis2020retrieval, borgeaud2022improving}. Subsequently, its retrieval module has been optimized through techniques such as query rewriting~\cite{zheng2023take, dai2023promptagator, chan2024rqrag}, passage reranking~\cite{glass2022re2g, yu2024rankrag}, and adaptive efficient retrieval strategies~\cite{asai2023self, jiang2023active, yao2023react}. In the generation module, recent efforts have focused on improving generation efficiency via prompt compression~\cite{jiang2024piperag, xu2024recomp, cheng2024xrag} and decoding optimization~\cite{merth2024superposition, jin2024ragcache, wang2025speculative}. These approaches primarily enhance individual RAG components in non-adversarial settings, whereas our research focuses on adversarial conditions (internal and external knowledge containing noise).

\textbf{Trustworthiness of LLMs in RAG.}
Ensuring the trustworthiness of LLMs in RAG systems has become a critical research area, particularly in adversarial environments. In scenarios where internal and external knowledge sources conflict, some studies align model responses with internal knowledge~\cite{pan2023risk, xu2023earth, chen2024universal, hong2024so, weller2022defending}, while others prioritize external knowledge~\cite{li2022large, gekhman2023trueteacher, zhou2023context}. 
Beyond these binary choices, recent efforts to improve the trustworthiness of LLM responses attempt to merge conflicting sources by using contrastive decoding~\cite{jin2024tug, shi2024trusting} or by integrating consistent portions of internal and external knowledge~\cite{xiang2024certifiably, wang2024astute, wei2025instructrag, zhou2025trustrag}.
However, these methods rely on carefully designed prompt templates to assess knowledge consistency. This matching strategy is sensitive to textual variations and lacks mechanisms for handling unanswerable questions. Previous work on enhancing LLM refusal capabilities in RAG systems relies on building expert knowledge bases~\cite{cao2023learn} or fine-tuning generative models~\cite{song2025measuring}. 
However, these methods often fail to integrate multiple knowledge sources and do not provide plug-and-play compatibility with black-box, model-based RAG systems.
Collectively, these works are orthogonal, typically targeting only one or two scenarios.